\documentclass{article}

\usepackage{nips13submit_e}
\nipsfinaltrue

\usepackage[utf8]{inputenc} 
\usepackage[T1]{fontenc}    
\usepackage{hyperref}       
\usepackage{url}            
\usepackage{booktabs}       
\usepackage{amsfonts}       
\usepackage{nicefrac}       
\usepackage{microtype}      

\usepackage{graphicx}
\usepackage{float}
\usepackage{multirow}
\usepackage{amsmath}

\graphicspath{{images/}}

\title{User Intent Classification using Memory Networks: A Comparative Analysis for a Limited Data Scenario}

\author{ Arjun Bhardwaj, Alexander Rudnicky\\
Carnegie Mellon University}

\begin{document} 

\maketitle

    

\section{Overview}

In this report, we provide a comparative analysis of different techniques for user intent classification towards the task of app recommendation. 
We analyse the performance of different models and architectures for multi-label classification over a dataset with a relative large number of classes and only a handful examples of each class.
We focus, in particular, on memory network architectures, and compare how well the different versions perform under the task constraints.
Since the classifier is meant to serve as a module in a practical dialog system, it needs to be able to work with limited training data and incorporate new data on the fly. 
We devise a 1-shot learning task to test the models under the above constraint.
We conclude that relatively simple versions of memory networks perform better than other approaches. Although, for tasks with very limited data, simple non-parametric methods perform comparably, without needing the extra training data.



\section{The Experimental Setup} \label{sec:firstpage}

\subsection{ The Data }

When people interact with their smart phones, they often have an “intent” – a high level task in mind. For eg, planning a dinner with friends. Just as often, there is not a single app that is able to meet the requirements, so people often pick and choose different functionalities from a number of apps in order to achieve their task. For eg, they might use Yelp to see reviews of nearby restaurant, use Google Maps to view locations, Uber to hail a cab and some messenger like Google Hangouts to communicate the plans to their friends.
 
\cite{Sun2016WeaklySU} describes a study in which the app usage of around 20 participants was monitored over a period of time, by a logger program running on their smartphone devices. From this, apps used in close temporal proximity to each other were identified. Given these sequences of apps, the participants were asked to provide a high level description of what task they were trying to perform / what goal they were trying to achieve, when they had used these apps. The users thus provided a description of the “high level intent” (HLI) that guided each of the specific app interactions. 
Samples from the resulting corpus can be found in Figure 1.

The study conducted by \cite{Sun2016AdaptingSD} collected analogous data by following a similar procedure. We, thus, merge compatible data from both these studies in our training corpus.

\begin{figure}[ht]
  \centering
  \label{tbl:table_q_aug_perf}
  \includegraphics[width=.5\textwidth]{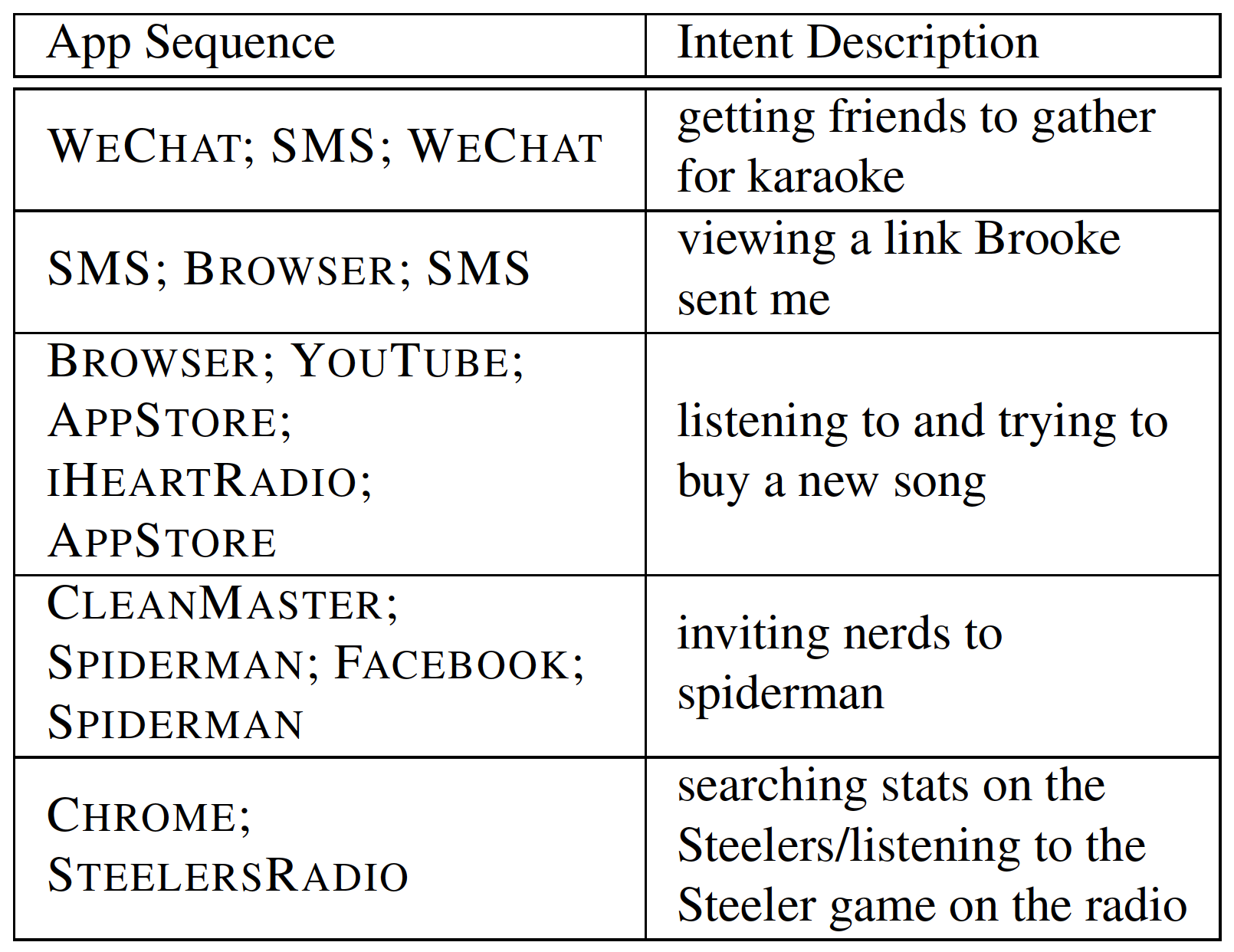}  
  \caption{App Sequences and the corresponding High Level User Intents they help realize}
\end{figure}

An observation to be made about the dataset is that the HLIs are abstract enough to make it non-trivial to infer the apps corresponding to a given HLI. Information that would have given the models an unfair advantage, like the HLI explicitly containing name of the app etc., is quite rare.
The situation is further complicated by the fact that the HLIs describe a high-level task that can be performed by employing a sequence of apps, and not just a single app in particular. This makes it difficult to use these HLIs to form a “profile” for each of the apps directly.

\subsection{Problem Formulation}

We thus have a corpus in which a natural language utterance - a HLI $hli_i$ - describes a task that can be performed using sequence of apps - $a_1^i,a_2^i ... a_m^i$.
A realistic formulation would have involved predicting the exact sequence apps corresponding to the given HLI, but given the limited size of the dataset (as described below), there would not have been nearly enough data to train such models.
To make the ML prediction task tractable, we relax the need to predict the exact sequence of apps, and instead try to simply predict the set of apps that would help realize a given HLI.
We thus treat the problem as a multi-label, multi-class classification problem, where we assign a set of app-id labels to each HLI.

\subsection{Task constraints}

The intent classification is meant to be in service of an intelligent smartphone assistant – HELPR \cite{Sun2016AnIA}. The model must recommend a set of apps to use, given a user’s utterance describing the task they wanted to perform. 
Given a recommendation, the user could let HELPR open the apps in the set (thereby indicating that the suggestion was indeed relevant), or alternatively select another relevant app from the list of apps on the device – thereby providing feedback in two different ways. 

The user could also provide a new high level intent and teach HELPR which app would be most relevant to the task. An example scenario would be: the user providing a natural language description of what the user would like to do with a new app that HELPR did not have in its database yet.

This usage scenario created an interesting set of requirements and constraints on the design of the recommendation system.
For one thing, it meant that the system was likely to be open ended, where the models should be flexible enough to incorporate new incoming data. This meant that online algorithms, or methods that did not involve very extensive training were a better fit.
Another aspect was that feedback was going to be available. The system should thus be able to make use of user’s feedback about its results and improve from this. Ideally we would like the model to be able to make use of both positive and negative feedback.

As another influencing factor, we had to decide how to present the results to the user. We could, for eg, decide that for any intent, we would show the top 3 or the top 5 app predictions, from the ranked list of relevant apps produced by the model. Alternatively, we could use a threshold based model in which the model itself controls the number of predicted relevant apps, based on the number of apps whose scores crossed an empirically determined threshold. Each of these decisions had implication for how the models were to be evaluated for accuracy, and how we could compare performance across models.

\subsection{Corpus Stats}

The corpus consists of 1590 sample points, and 196 apps in all. We randomly split the corpus in ration of 60:15:25 to form the training, validation and test set, leaving us with 954, 238 and 398 samples in the respective sets. The corpus is heavily skewed, as can be seen from the Fig 2.

\begin{figure}[ht]
  \centering
  \label{CS100}
  \includegraphics[width=.5\textwidth]{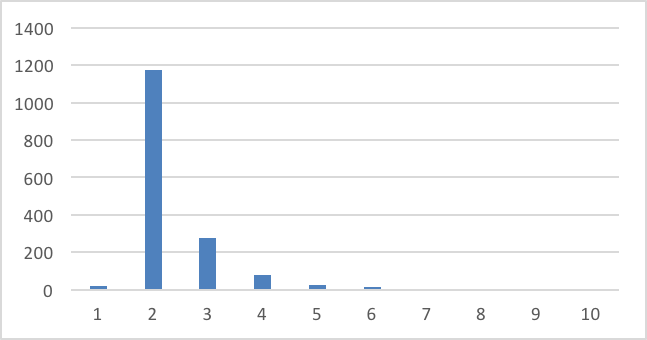}
  \caption{The x axis represents the number of distinct apps used to realize a particular high level intent. The y axis represents the number of high level intents in the entire corpus. Thus, most high level intents require 2 apps, though some require more.
}
\end{figure}

The amount data we have for modelling each app class, is presented in Table below.

\begin{table}[ht]
\centering
\label{tbld:CS101}
\begin{tabular}{|c|c|}
\hline
Number of Apps  &Number of HLI\\ \hline
176	&[1, 49)  \\ \hline
13	&[49, 97)  \\ \hline
2	&[97, 145) \\ \hline
0	&[145, 193) \\ \hline
3	&[193, 241) \\ \hline
\end{tabular}
\caption{Histogram data for number of HLIs for each app}
\end{table}

This goes to show how there are some apps, which are more popular and used more frequently, across different high level intents. These could be apps like Google Chrome, Facebook etc. Let’s look a bit more closely at the above stat, while zooming in on the initial part of the trend.

\begin{figure}[ht]
  \centering
  \label{CS102}
  \includegraphics[width=.5\textwidth]{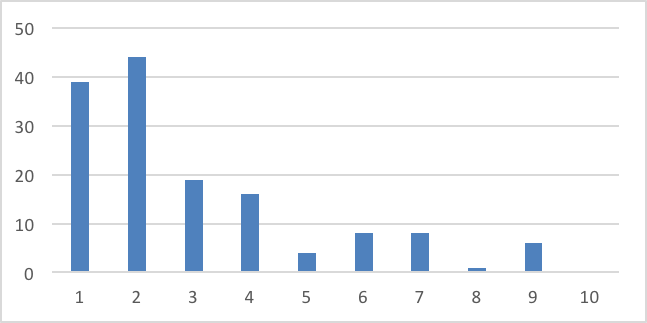}
  \caption{The x axis represents the number of distinct apps used to realize a particular high level intent. The y axis represents the number of high level intents in the entire corpus. Thus, most high level intents require 2 apps, though some require more.
}
\end{figure}

Thus, in the corpus, we do have around 40 apps, for which we have only one HLI describing their functionality. Thus, if these were never encountered in the training set, it would be impossible to correctly predict them in the test set. However, we let these populate all the corpus partitions, without any restriction. The reason being that it better approximates the real world setting, in which the user may communicate that they need to perform a task for which though there is an app, but HELPR doesn’t know about it.

\subsection{Trivial Baselines, Bounds and Metrics}

Given the nature of the dataset, we wanted to test how a simple majority class baseline would perform. In this case, the model simply returns a list of apps ranked according to their frequency of occurrence in the training corpus. 

For this method, as for some of the other methods we discuss later, we combined the training and the validation sets to form a joint training set, while the results are from the test set.
The reason for doing this was to mimic practical conditions. In real life, given scarce data, models that require validation datasets would end up with lesser data for training. Thus, by this strategy, this trade-off manifests itself in the model’s performance metrics.
The results are presented in Table\ref{tbl:MajBas}.

\begin{table}[ht]
\centering
\caption{Performance (in \%) for Majority Baseline approach}
\label{tbl:MajBas}
\resizebox{\textwidth}{!}{
\begin{tabular}{|c|c|c|c|c|c|c|c|c|c|c|}
\hline
&top-1    & top-2 & top-3 & top-4 & top-5 & top-6 & top-7 & top-8 & top-9 & top-10   \\ \hline
coverage & 12.71 & 25.01 & 30.82 & 37.02 & 42.45 & 45.41 & 48.98 & 51.51 & 54.23  & 55.88 \\ \hline
purity   & 28.89 & 28.14 & 23.28 & 21.04 & 19.4  & 17.42 & 16.08 & 14.98 & 13.99  & 12.94 \\ \hline
\end{tabular}}
\end{table}

At this point, we define two metrics: purity and coverage.
We adopt a model output format where we take the top-n results from the ranked list returned by the model, for a particular HLI. Let this be the list of the predicted app, as opposed to the list of actual apps for the HLI, which we note from the corpus.
For determining purity, we look at just how “pure” the list of predicted apps is – i.e. what percentage of the apps in the predicted list were in the list of actual apps.
For coverage, we look at what percentage of actual apps were “covered” - i.e. what percentage of actual apps made it into the predicted apps list.

We use results in this format to compare across models.
This was adopted over a threshold based scheme where a variable number of apps made it into the list output by a model. The reason for doing so was that setting a threshold would have required a separate empirical training on an additional dataset in many of the models. But more than that, it was a practical design decision, where the HELPR agent would just always output something like 3 or 5 apps in its results section.

We pay the most attention to the top-3 scores in our discussions. Sometimes for comparison, we use the top-5 and top-1 scores too. The bold figures in the score tables point to this fact only.

Next, we try to determine what a realistic maximum level of performance for the dataset would be. For doing this, we assumed a semi-perfect model that detected any app which had occurred in the training set i.e. for an example in the test set, the predicted app list was composed of all the apps in the actual app list which had occurred in the training set. In case the resulting list consisted of less than n results in the top-n setting, the remaining slots were filled using the top members of the majority class list described above. The results are presented in Table \ref{tbl:PerMod}

\begin{table}[ht]
\centering
\caption{Performance (in \%) for Majority Baseline approach}
\label{tbl:PerMod}
\resizebox{\textwidth}{!}{
\begin{tabular}{|c|c|c|c|c|c|c|c|c|c|c|}
\hline
& top-1    & top-2 & top-3 & top-4 & top-5 & top-6 & top-7 & top-8 & top-9 & top-10       \\ \hline
coverage & 45.03 & 88.63 & 95.31 & 96.78 & 97.4  & 97.6  & 97.6  & 97.6  & 97.6   & 97.6  \\ \hline
purity   & 98.24 & 97.11 & 72.19 & 55.84 & 45.33 & 37.98 & 32.56 & 28.49 & 25.32  & 22.79 \\ \hline
\end{tabular}}
\end{table}

One can see here that for higher values of n, we get a better coverage. It almost saturates, starting from 3 onwards. This is probably because an adequately large number of HLIs in the dataset are realized by just $\leq$3 apps.

These numbers about the purity or coverage of this perfect model, provide a helpful context when evaluating the performance of our models below, acting as a good/practical upper bound.

\section{Baseline Models for Intent Classification}

\subsection{The Word2Vec Nearest Neighbour Model}

In this model, we used the pre-trained word2vec embeddings to produce a representation for each HLI. These were used in a nearest neighbour classifier. 

\paragraph{A word about the word2vec embeddings}: These embedding are used quite pervasively in NLP. We try to motivate the intuition behind them here. For further details, the reader is referred to \cite{Mikolov2013EfficientEO}. The idea was to feed a one hot representation of a word to a set of neural layers. These layers were broader at the ends (near the input and the output nodes) and thinner in the middle. The output desired was a vector which had 1’s at the indices which stood for the words that happened to occur in the same context as the input word (e.g., words which happened to occur within n=2 word window from the input word). 
Because of the structural constraint, the network tries to form representations near the middle layers that assign similar activations to words that share similar properties – these properties could be syntactic, semantic etc. 
When the model is trained, the encoder part of the network – the one consisting of layers leading up till the middle layer – is used as a general purpose embedding network to convert words to their vector representations.

In our model, we form the sentence vector representation for each of the HLIs by averaging the word vector representations for their constituent words. 
Next we form a profile for each app - a set of vectors defining the functionalities of this app. These consist of the sentence vectors for all the HLIs that included this app in the set of apps that were used to realize it.

When we have a new query, we compute scores for each app. We first find the closest vector to the query, from amongst the set of HLI vectors for this app. The score for the app is the inverse of the Euclidean distance between the query and this vector.

The model’s output is a list of apps, ranked according to this score.

The model is simple and works well. The decision to use Euclidean distance (instead of, say, cosine similarity etc.) and min of distance (as opposed to an average etc.) was empirically motivated. The accuracy was found to be the highest for these, based on a model that was trained on the training partition and tested on the validation set.

\textbf{Advantages:}
A big advantage of the model is that it can incorporate new data easily. When the user adds a new description for a app, the description can be turned to a vector on the fly and added to the app’s HLI set.

\textbf{Disadvantages:}
When there are millions of entries, this model would not scale. Though there are many well studies optimizations that could be made. However, this was not a big concern for the current dataset.

\textbf{Results:}
For the final testing, we included the validation set into the training set, and tested on the training partition that had been kept aside for this purpose. The results are documented in Table \ref{tbl:w2vres}.

\begin{table}[ht]
\centering
\caption{Performance (in \%) for Word2vec Nearest Neighbour Model}
\label{tbl:w2vres}
\resizebox{\textwidth}{!}{
\begin{tabular}{|c|c|c|c|c|c|c|c|c|c|c|}
\hline
& top-1    & top-2 & top-3 & top-4 & top-5 & top-6 & top-7 & top-8 & top-9 & top-10       \\ \hline
coverage & 29.96 & 62.92 & 75.36 & 79.58 & 81.87 & 84.13 & 86.08 & 87.09 & 88.4   & 89.02  \\ \hline
purity   & 65.33 & 68.84 & 55.95 & 44.79 & 37.14 & 31.91 & 28.07 & 24.91 & 22.53  & 20.45 \\ \hline
\end{tabular}}
\end{table}

We found this to be a very hard model to beat. Though simple this model worked well under the constraints of the experimental conditions.

\subsection{Skip Thoughts Nearest Neighbour Model}

In this version of the above approach, instead of using word2vec embeddings, we used skip thoughts encoder to form the vectors for the HLIs. The hope was that these would be able to do a better job at forming a vector representation capturing the entire meaning of the sentence, since this model had that as its direct aim (as compared to word2vec which aimed to form a vector form for words only).

\begin{figure}[ht]
  \centering
  \label{CS1}
  \includegraphics[width=.8\textwidth]{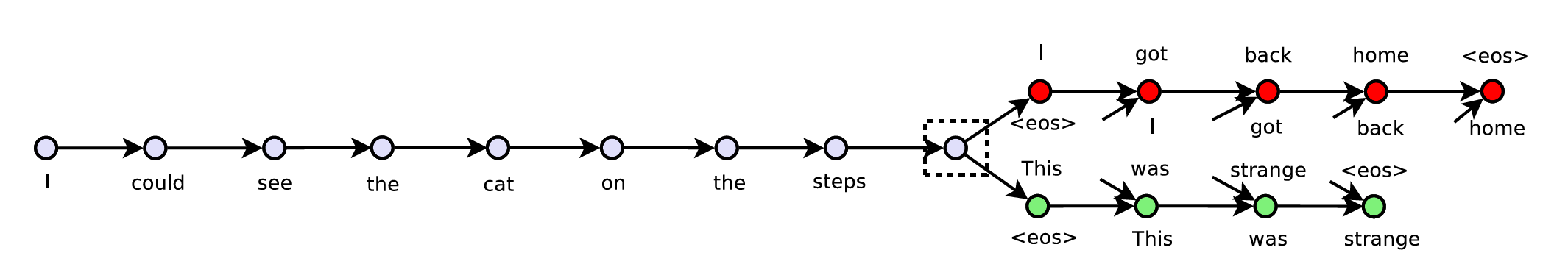}

\end{figure}

We try to motivate the intuition behind the model.
The model consisted of a RNN (for eg, a LSTM) that sequentially reads the words of a sentence. The final output of this encoder LSTM then feeds into a decoder LSTM as its initial state. Given this vector summary for the sentence, this LSTM then tries to generate the preceding sentence, in one case, and the succeeding sentence, in another case.
The idea is similar in a way to word2vec. There the assumption was that the meaning of a word is captured by the surrounding words, while here it seems to be that the meaning of a sentence is captured by its surrounding sentence.
For details, the reader is referred to \cite{Kiros2015SkipThoughtV}.

We used the pre-trained model provided by the authors, which was trained on a set of books.
The rest of the nearest neighbour set up was the same. Results of the approach are presented in \ref{tbl:skthres}:

\begin{table}[ht]
\centering
\caption{Performance (in \%) for Skip Thought Nearest Neighbour Model}
\label{tbl:skthres}
\resizebox{\textwidth}{!}{
\begin{tabular}{|c|c|c|c|c|c|c|c|c|c|c|}
\hline
& top-1    & top-2 & top-3 & top-4 & top-5 & top-6 & top-7 & top-8 & top-9 & top-10       \\ \hline
coverage & 30.21 & 61.5  & 69.76 & 75.82 & 78.47 & 80.45 & 81.38 & 83.58 & 84.65 & 85.67 \\ \hline
purity   & 65.08 & 67.09 & 52.18 & 42.9  & 35.63 & 30.49 & 26.49 & 23.9  & 21.55 & 19.67 \\ \hline
\end{tabular}}
\end{table}

As we see from the top-1 and top-3 scores, there wasn’t much of an improvement. 
One explanation for this might be that the model was trained on a set of storybooks, which was unable to perform too well on the casual modern speech that comprised the HLIs.

\subsection{Siamese LSTM Nearest Neighbour Model}

Hoping to model the semantic similarity between two sentences directly, we used the model proposed by \cite{Mueller2016SiameseRA}.

The model is based on an intuitive idea. Unlike the previous two models, for which semantic similarity was a more like a side-effect of the actual training target, this model is trained explicitly to predict the semantic similarity of two sentences. The training data for this came from SemEval semantic similarity tasks and the authors took some more processing steps to ensure that the model could generalize beyond the limited sample.

\begin{figure}[ht]
  \centering
  \label{SiaDiag}
  \includegraphics[width=.8\textwidth]{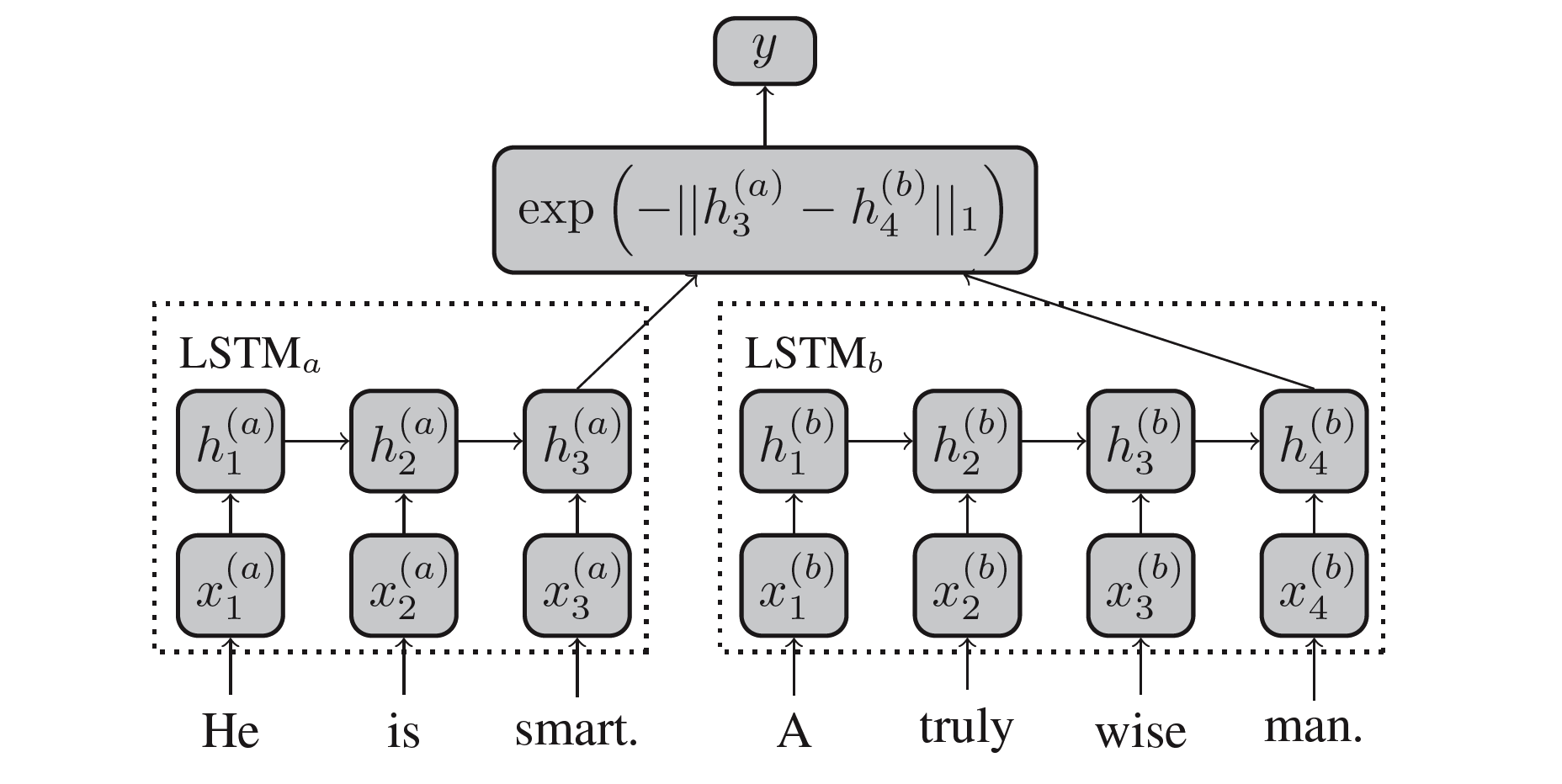}

\end{figure}

The main idea is as follows: a LSTM reads the sequence of words in one sentence, and the final output from the LSTM is taken as the sentence vector representation. The same LSTM also operates on the other sentence, turning it into a vector too. A final, neural layer then looks at a vector capturing the differences between these vectors and predicts a score between 1-5 (5 being the most semantically similar).

We used a pretrained version of model to directly calculate the semantic similarity score between a test query and the set of HLIs that form the profile of an app. The ultimate score for an app was then the maximum similarity score between the query and any of its HLIs. Results obtained are documented in the Table \ref{tbl:siares}.

\begin{table}[ht]
\centering
\caption{Performance (in \%) for Siamese LSTM based Nearest Neighbour Model}
\label{tbl:siares}
\resizebox{\textwidth}{!}{
\begin{tabular}{|c|c|c|c|c|c|c|c|c|c|c|}
\hline
& top-1    & top-2 & top-3 & top-4 & top-5 & top-6 & top-7 & top-8 & top-9 & top-10       \\ \hline
coverage & 26.67 & 56.71 & 66.7  & 70.71 & 72.73 & 75.3  & 77.18 & 78.08 & 79.26 & 80.91 \\ \hline
purity   & 57.79 & 61.18 & 49.08 & 39.51 & 32.66 & 28.31 & 24.95 & 22.14 & 20.02 & 18.39 \\ \hline
\end{tabular}}
\end{table}

The results were interesting, though not as encouraging. While the Top-1 score went up, the top-3 score went down.
Further investigation is needed to be able to discern the exact reason for the same. However, one suspicion became stronger that perhaps our very assumption of using semantic similarity to predict apps is misplaced, or at least not relevant. The kind of semantic similarity needed for app prediction might not be the same as the one described in the SemEval tasks.

The motivation for using this model was as follows.
Knowing that messenger is one of the apps used for the HLI “inviting nerds to spiderman”, we wanted this HLI to be sematically similarity to another HLI “coordinating with friends about rowing practice”, to infer that the latter HLI can use messenger too … 
But this model likely doesn’t see these sentences as being adequately similar, given that this model is trained in a way as to think that sentences describing the same event, and the same actors and objects are what should be similar. This is clearly not valid for this pair of intents, just like many others.

\subsection{Feedforward neural network (FFNN)}\label{sec:FFNN}

Next, we tried out a simple FFNN model. The model was trained on the training set of 954 points, and we used the validation set for early stopping.

We varied the architecture in terms of the number of layers the network must have (1,2,3), the number of neurons each layer must support (50,100,300) and the choice of the loss function (cross entropy loss with softmax output layer or rmse with sigmoid output layer ).
Intuitively too, there are merits in the use of a sigmoid output layer, rather than a softmax layer since our task is one of multi-label classification, while softmax function is best suited for single label classification since it forces there to be a relatively sharp peak in the activation of only one node in the output layer.

\begin{figure}[ht]
  \centering
  \caption{Purity on the Training (Red) and Validation (Blue) Sets for the FFNN Model.}
  \label{FFNNF1}
  \includegraphics[width=.5\textwidth]{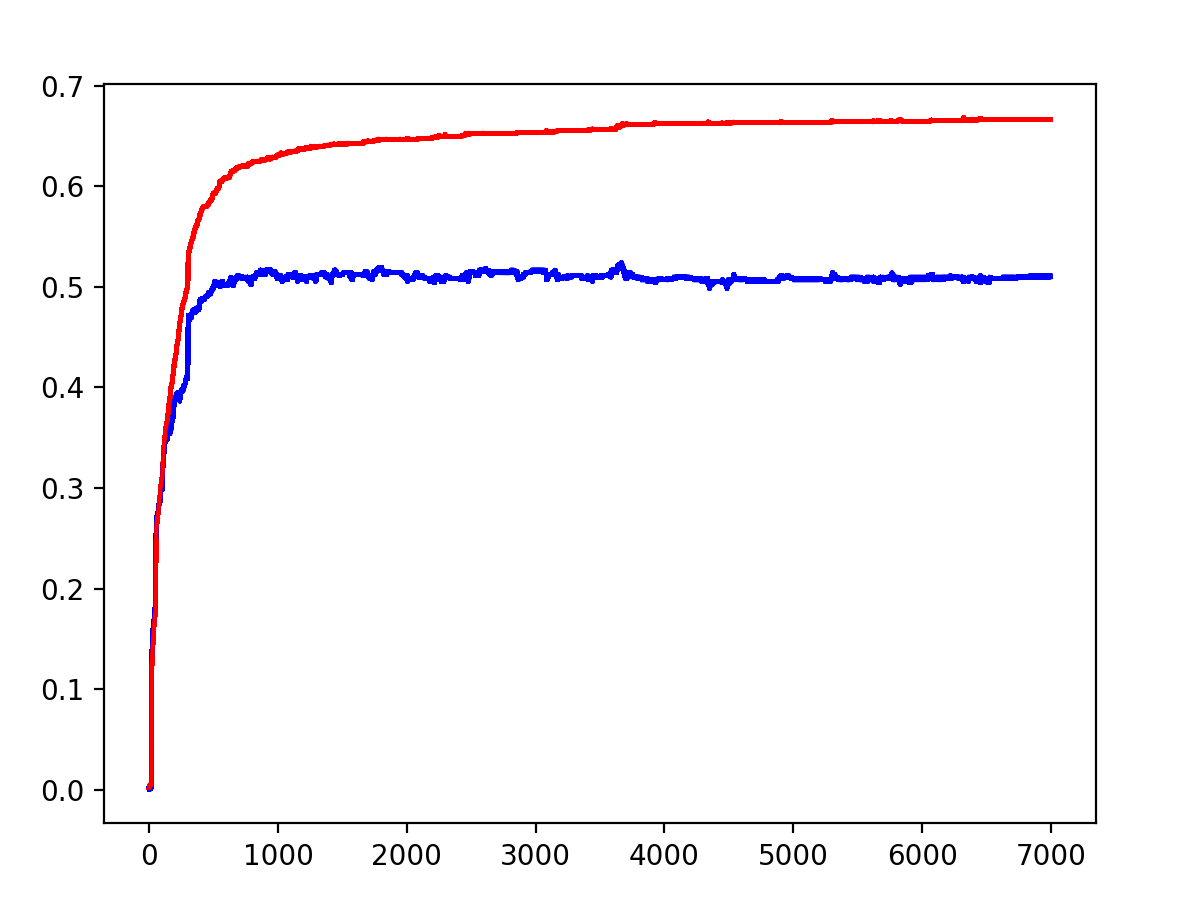}
\end{figure}

We emperically found that that a neural network with 2 simoid output layers of dimension 100 each performed the best. Figure \ref{FFNNF1} presents the top-3 accuracy on the validation dataset.

The red line is for the accuracy on the training partition while the blue line is for the validation partition.
We used multiple random starts and Adam as the optimizer, which helped prevented overfitting.
The results are tabulated in Table \ref{tbl:ffnnres}.

\begin{table}[ht]
\centering
\caption{Performance (in \%) for FFNN Model}
\label{tbl:ffnnres}.
\begin{tabular}{|c|c|c|c|c|c|c|c|c|c|c|}
\hline
& top-1    & top-2 & top-3 & top-5      \\ \hline
purity & 81.4  & 66.2  & 50.8  & 33.1 \\ \hline
\end{tabular}
\end{table}

The top3 result wasn’t able to outperform the nearest neighbor approach, but we curiously see that the top-1 results is much better.

We also tried another paradigm, in which we used a threshold of 0.5 to decide if an app should make it to the output predicted app list.
With this, the purity went to 71.9 but the coverage stood at 58.7

\section{Memory Network based approaches}

We now describe approaches to building the classifier that involve memory network architectures.
The motivation behind using memory networks \cite{Sukhbaatar2015EndToEndMN} was to combine the beneficial aspects of nearest neighbor and neural approaches. Several recent papers have proposed the use of memory networks for the few-shot learning scenarios. The memory network is trained to make use of the labelled examples in its memory, to predict the classification labels of the test cases. In such a setting, the set of memories can easily be expanded as new labelled samples come in, thus making this model a natural match for our task.

We first briefly describe relevant memory network architectures. We then describe how we adapted and applied these architectures to our task and compare the performance of these models.

\subsection{End to End Memory networks}\label{sec:e2eMem}

In the paper (Sukhbaatar et al.), the authors showed how a memory augmented model could be applied to the task of machine comprehension.
Viewing from a high level, the memory network is basically a way of augmenting an input query vector with “relevant” information contained in the memories.
The above process can be referred to as a "hop".
We now describe this further.

For any hop $k$, the input query vector - $q^k$ - is transformed, by a single layer feedforward neural layer (or as the paper describes it, by an embedding matrix $B^k$), into $u^k$.
Information relevant to the task is formed into a set of vectors $x_i$ which populate the memory of the network.
Each of the memory vectors - $x_i$ - are also transformed by embedding matrices $A^k$ and $C^k$, into - $m_i^k$ and $c_i^k$. We refer to these, here, as the x-form and y-form of the memory vector $x_i$, respectively.
In a hop, the first step, intuitively, involves calculating the relevance of each of the memories w.r.t. the query. 
This relevance between $u^k$ and any $m_i^k$ is captured by the following expression : $(x_i^k)^Tm_i^k$.
We thus obtain a set of scores, which then undergo a softmax normalization, resulting in a vector which has a dimension equal to the number of memories.
The output from the memory, $o^k$, is then formed by summing each of the memory y-forms, weighed by these normalized relevance scores.
The final response of the hop is computed by adding the relevant information retrieved from the memory to the input vector, as follows : $u^{k+1}=o^k+u^k$.
This chain of steps forms a single “hop”.

At any hop $k$, the response $u^k$ can then serve as the query vector $q^{k+1}$ for the next hop. To limit the number of parameters, the paper discusses weight sharing schemes by which embedding matrices across hops can be constrained to be the same.
To generate the final prediction, for the model consisting of $K$ hops, $u^k$ from the final hop is passed through another embedding matrix $W$ to produce the prediction vector $a$.

In general, a larger number of hops means that more information can incorporated into the query, in more complex ways. The number of hops thus can have a direct implication on just how complex a reasoning can be performed by the network.

\subsection{Matching Networks for One shot learning}\label{sec:MatNet}

The model proposed by \cite{Vinyals2016MatchingNF} can be seen as an adaptation of the above memory networks towards the task of performing classification in general, and for few-shot learning in particular.

The target of the proposed model is single-label, multi-class classification - to map from a support set of example pairs consisting of an input $x_i$, and its corresponding output - a label distribution $y_i$: $S = {(x_i,y_i)_{i=1}^k}$ to a classifier $c_S$ which takes an input $\hat{x}$ and produces a probability distribution over labels as the output. In the simplest case, this takes the following form:
\begin{equation}
    P(\hat{y}|\hat{x},S) = \Sigma_{i=1}^k a(\hat{x},x_i)y_i
\end{equation}
where $a(\hat{x},x_i)$ is an attention mechanism and can take the following form, with $c$ being the cosine distance:
\begin{equation}
    a(\hat{x},x_i) = e^{c(f(\hat{x}),g(x_i))} / \Sigma_{j=1}^k e^{c(f(\hat{x}),g(x_j))}
\end{equation}

Some readers may note the similarity of the above procedure to case based reasoning methodologies.
The functions $f,g$ can be something as simple as feedforward neural networks acting upon their respective inputs (potentially with $f=g$).

The paper introduced some other important strategies to help the model perform better. The function $f$ responsible for forming a representation for the input $\hat{x}$ can be built in a way that it uses the support set $S$. The paper demonstrates how this can be done using a LSTM.

Even when training the model, the batches can be formed in a way that encourages the model to learn how to selectively use the information contained in $S$, to make the test prediction.
To do this, they sample a set of labels from the entire set of labels. A batch $L$ is prepared containing training examples which have labels in this set of labels.
Support sets and Training batches are then sampled from $L$.
For details regarding the optimization target, the reader is referred to the original paper.

\subsection{Models and Results}

\subsubsection{Query Augmentation Memory Networks}\label{sec:QAMNTest}

We first evaluated the performance of variations of the \cite{Sukhbaatar2015EndToEndMN} model (discussed in \ref{sec:e2eMem}) when applied directly to our task.

We were interested in comparing the performance of this approach to the results we had obtained earlier using the feedforward neural net model.
While in the earlier model a feedforward net transformed the test query HLI $q$ into the output prediction vector $a$ directly, here we are augmenting the query vector with relevant information from a support set consisting of the training examples, before passing it to a similar feedforward net.

We now describe the setup in more detail. 
The user's utterance is tokenized and converted to a query vector $q$ using the same procedure as the one employed in the word2vec model described earlier.
$q$ is then fed to a memory network consisting of $K$ hops, resulting in the vector $u^K$.
This is then fed to a single layer of neurons with softmax activation to produce the prediction vector $a$.
Each element in $a$ is interpreted as a score for the app corresponding to that index.
We refer to this architecture, here, as the query augmentation memory network.

We empirically repeatedly found that the use of softmax activation in the final layer yielded a higher performance that the use of sigmoids. The reason could be the fact that in softmax activation ties the outputs of the neural nodes together, in a way that an activation for one node leads to a corresponding de-activation of all others. In comparison, for sigmoid activation, the neural nodes are independent, and since the target vector has only a handful of indices of the 196 with a non-zero value, the overall learning is dominated by the need to make the output 0.

In our experiments we  tested the performance for variations of the architecture. For eg, we varied the number of hops and compared the performance for models with 1 and 3 hops.
Since the training data is limited, to reduce the number of parameters, in a variation, we set the embedding matrix $A^k$ to be the same as $B^k$. This was also done to enable a comparison between this model and the matching network model described in \ref{sec:MNTest}, which employs a similar weight sharing scheme.
Finally, we explore the effect of introducing non-linearity. 
In a variation, we set $m_j^k = tanh(x_j^T A^k + Bias_A^k)$ (where $Bias_A^k$ is a bias vector) instead of simply setting $m_j^k = x_j^T A^k$. 
This transformation can be seen as passing $x_j$ through a single layer feedforward network.
$c_j^k$ and $a^k$ undergo a similar transformation as well.

When training, we sample a batch $\mathcal{B}$ from the training set $\mathcal{T}$. The HLIs corresponding to the samples in $\mathcal{T}-\mathcal{B}$, then populate the set of memory vectors $x_j$, for this batch.
We used the Adam optimiser with a RMSE loss function.
We employed early stopping with multiple random starts.

We present the performance of the models on the validation dataset in Table \ref{tbl:table_q_aug_perf}.
For each variation of the model, we present the top-1 and the top-3 purity metrics to represent the model's performance.

\begin{table}[ht]
\centering
\caption{Accuracy (in \%) of different models on the validation dataset}
\label{my-label}
\resizebox{\textwidth}{!}{
\begin{tabular}{|c|c|c|c|c|c|c|c|c|}
\hline
&\multicolumn{4}{|c}{with non-linearity} & \multicolumn{4}{|c|}{without non-linearity}\\ \hline
&\multicolumn{2}{|c|}{with weight sharing} & \multicolumn{2}{c|}{no weight sharing} &\multicolumn{2}{c|}{with weight sharing} & \multicolumn{2}{|c|}{no weight sharing}\\ \hline
        & 1-hop & 3-hop & 1-hop & 3-hop & 1-hop & 3-hop & 1-hop & 3-hop \\ \hline\hline

top-1 purity    & 31.51 & 32.35 & 31.93 & 45.8 & 17.23 & 7.14  & 11.76 & 2.10  \\ \hline
top-3 purity    & 10.92 & 11.34 & 12.74 & 21.71 & 7.56 & 2.94  & 5.32 & 1.10  \\ \hline
\end{tabular}}
\end{table}

The results show that the architecture is likely not an optimal choice for this task.
We notice that the presence of non-linearity increases the performance.
We see that increasing the number of hops does not increase the performance in all cases. In the linear cases, increasing the number of hops in fact degrades performance.

None of the model variations are able to perform better than the feedforward neural network.
This remains the case even when we substitute the single neural layer transforming $u_k$ to $a$, for a double neural layer, more closely resembling the best performing result from \ref{sec:FFNN}.
One explanation for this might be that perhaps the query augmentation approach is unable to work well because the vectors involved are dense word2vec vectors, as opposed to simple few-hot or bag of words style vectors that were used in the original paper. When working with these simpler representations, simple operations like addition preserve information, while the same can not be said for the dense vectors since they encode information in a more complex manner.
More experiments need to be carried out to explore this further.

The best result obtained was for the version with non-linearity, no shared weights and 3-hops. Increasing the number of hops beyond 3 did not lead to a further increase in performance.

\subsubsection{Matching networks}\label{sec:MNTest}

We now discuss the performance of models using variations of the matching network architecture discussed in \ref{sec:MatNet}.

The model achieved state of the art results on few-shot learning tasks for multi-class, single-label classification. One of the most challenging task involved a 1-shot 20-class classification, where correct label had to be assigned out of a set of 20 novel labels never before seen during training, with only 1 labelled example in the support set. 
One may note at this stage that our task is one of selecting a variable number of labels from a set of 196 labels. Furthermore, while they trained the model on a much larger set before testing it with the novel classes, we have data only to the order of a 1000 points.
On account of the data limitations, we did not implement $f$ as a LSTM based function. We used a single layer neural network with 300 tanh activation units.

We also experiment with an architecture which can be seen as a hybrid of the query augmentation architecture with the matching network architecture.
Assume that the network needs to generate the prediction / output vector $a$ for an input query $q$. Assume that the model has a memory component consisting of $K$ hops, with the set $S=\{(x_j,y_j)\}$.
The initial part of the architecture is a memory augmentation memory model, consisting of $K-1$ hops, while the final hop is a matching network.

In terms of the query augmentation model, the $m_j^k$ and $c_j^k$ are formed by a non-linear transformation of $x_j$, as described in \ref{sec:QAMNTest}.
The initial query $q=q^1$ is thus passed through $K-1$ hops to form $u^{k-1}$.
Thereafter, in the final hop, $u^{k-1}$ serves as the input $\hat{x}$ for a matching network with support set $S$.
The matching network thus generates the prediction $a$ taking an augmented form of the query as its input.
For $K=1$, the hybrid architecture reduces to a simple matching network.

We further form an alternate version of the model in which the weight matrices and bias vectors of the tanh neural network layer responsible for transforming $x_j^k$ to $m_j^k$ ($A^k$ and $Bias_A^k$) and $q^k$ to $u^k$ ($B^k$ and $Bias_B^k$) is constrained to be the same.
Conceptually, this can be seen as casting the respective vectors, $x_j^k$ and $q^k$, into a shared space where the relevance between the two can be better measured by a cosine similarity metric.
This further reduces the number of free parameters that need to be estimated.

We trained the model with Adam optimizer using multiple random starts and early stopping.
The batches for training were prepared as follows. A batch $\mathcal{B}$ is sampled from the training set $\mathcal{T}$. The support set making up the memories of the network for this batch consists of $\mathcal{T}-\mathcal{B}$.
A batch size of 32 was used.
We present the results in Table \ref{matching_net_results}. 

\begin{table}[ht]
\centering
\caption{Accuracy (in \%) on the validation dataset}
\label{matching_net_results}
\begin{tabular}{|c|c|c|c|c|}
\hline
             & \multicolumn{2}{c|}{without shared weights} & \multicolumn{2}{c|}{with shared weights} \\ \hline
             & 1-hop                & 3-hop                & 1-hop               & 3-hop              \\ \hline\hline
top-1 purity & 70.17                & 59.24                & 83.13                  & 40.76              \\ \hline
top-3 purity & 44.12                & 33.47                & 57.14                  & 22.41              \\ \hline
\end{tabular}
\end{table}

The results are promising. 
The best performing model is one with shared weights and a single hop. 
It is likely that this is on account of the fact that this model had the lowest number of free parameters whose values needed to be estimated from the limited training set.
The model performs better than both the word2vec nearest neighbour approach and the feedforward neural network approach.
We empirically found that increasing the number of nodes in $f$, or realizing $f$ as a double layer feedforward neural network did not improve the performance further.
We present the performance of this model on the held-out test partition in Table \ref{best_model_test_set_perf}

\begin{table}[ht]
\centering
\caption{Performance (in \%) of the best model on the test partition}
\label{best_model_test_set_perf}
\resizebox{\textwidth}{!}{
\begin{tabular}{|c|c|c|c|c|c|c|c|c|c|c|}
\hline
& top-1    & top-2 & top-3 & top-4 & top-5 & top-6 & top-7 & top-8 & top-9 & top-10    \\ \hline
purity & 80.90 & 72.49 & 56.28 & 45.48 & 37.24 & 31.83 & 27.78 & 24.78 & 22.31  & 20.20 \\ \hline
coverage   & 37.25 & 66.84 & 75.91 & 80.75 & 81.98 & 83.60 & 84.94 & 86.33 & 87.36  & 87.84 \\ \hline
\end{tabular}}
\end{table}

\subsubsection{Testing 1-shot performance on unseen apps}

The best performing model so far is the memory network described above.
Several prior works have established the efficacy of the architecture on few-shot learning tasks.
However, these models do require some training. 
They need to be taught how to use the support set, before they can use the support set labelled examples to correctly label test examples with labels that the model never encountered during training.
In comparison, the word2vec nearest neighbour model \ref{sec:FFNN} is a non-parametric method that provides competent performance without needing any training.

We devise an experiment to test how accurately the two models predict unseen apps for test HLIs using only a single example from the support set.
However, the nature of our task is one of multi-label classification, and this makes it very difficult to create reasonably sized partitions of the corpus which have no apps in common, since apps widely co-occur with other apps.
Instead, we evaluate the models on a synthetic single-label classification task based on the dataset.

\begin{figure}[ht]
  \centering
  \caption{Corpus subsets}
  \label{1_shot_split}
  \includegraphics[width=.7\textwidth]{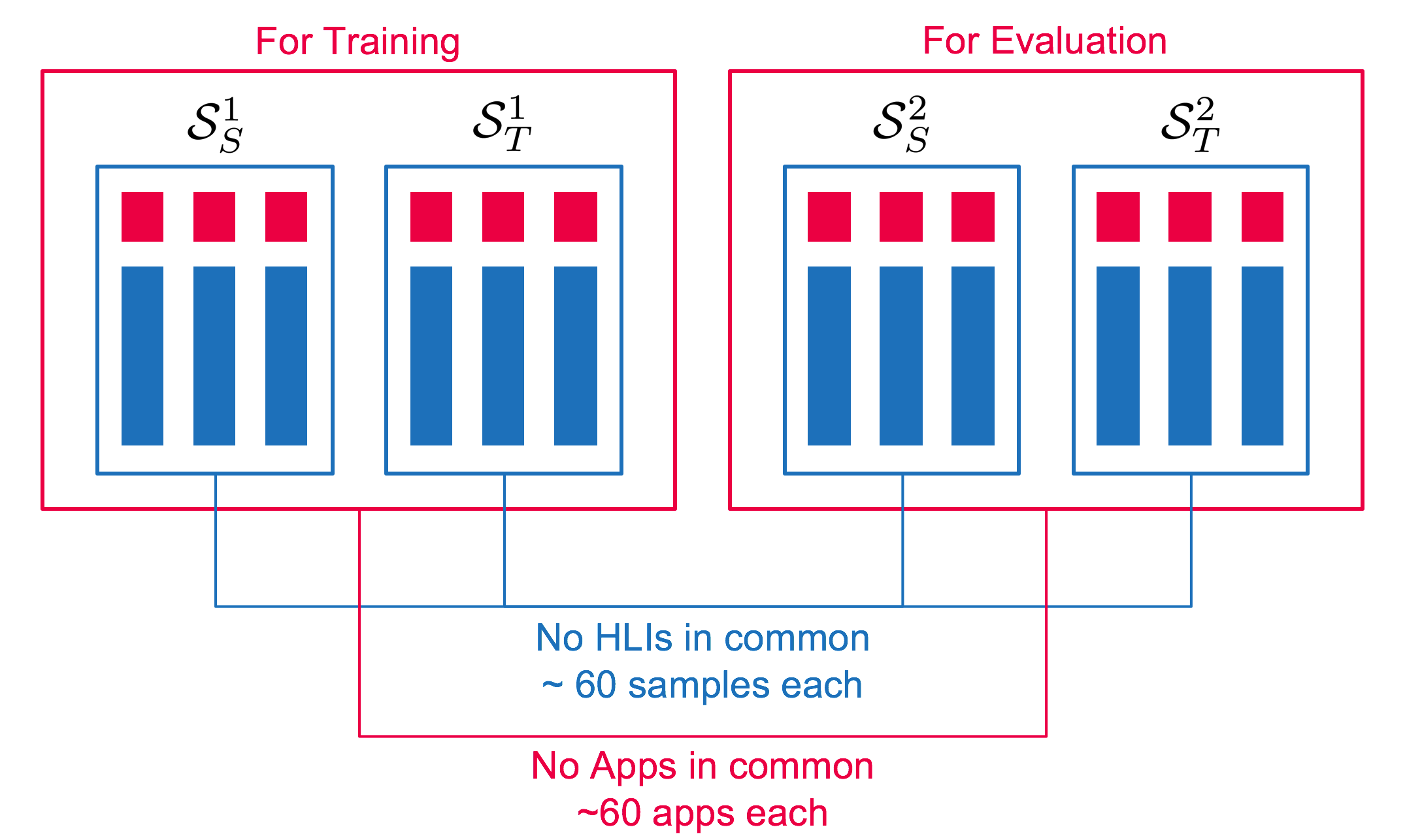}
\end{figure}

Let $\mathcal{S}=\{(h_j, A_j)\}$ represent the entire corpus, where $h_j$ is any HLI and $A_j=\{a_{j_1} ... a_{j_m}\}$ is the set of apps corresponding to it.
Let $L$ be the set of all app labels in the dataset.
We divide $L$ into two disjoint subsets $L^1,L^2$, such that each app in $L^1,L^2$ has more than 2 HLIs in the entire corpus. 
$\mathcal{S}^1=\{(h_j^1, a_j^1)\}$ and $\mathcal{S}^2=\{(h_j^2, a_j^2)\}$ are derived from $\mathcal{S}$, where $a_j^1\in L^1,a_j^2\in L^2$. 
We form support sets $\mathcal{S}^1_S\subseteq \mathcal{S}^1, \mathcal{S}^2_S\subseteq \mathcal{S}^2$ and training/test sets $\mathcal{S}^1_T\subseteq \mathcal{S}^1$, $\mathcal{S}^2_T\subseteq \mathcal{S}^2$.
$\mathcal{S}^1$ will be used exclusively to train the memory network, while $\mathcal{S}^2$ will be used for evaluating the performance of the models.
The sets are illustrated in Figure \ref{1_shot_split}.

Essentially, we decompose a multi-label example $(h_j,\{a_{j_1}...a_{j_m}\})$ into multiple single-label samples $(h_j,a_{j_1})...(h_j,a_{j_m})$.
Any of the single-label samples with app labels not belonging to $L^1$ (for $\mathcal{S}^1_S$,$\mathcal{S}^1_T$) or $L^2$ (for $\mathcal{S}^2_S$, $\mathcal{S}^2_T$)  are discarded.
The remaining samples are are assigned together to one of the 4 sets $\mathcal{S}^1_S,\mathcal{S}^1_T,\mathcal{S}^2_S,\mathcal{S}^2_T$.
To prevent unintended knowledge transfer, we further ensure that $\mathcal{S}^1$ and $\mathcal{S}^2$ share no HLIs. Similarly for $\mathcal{S}^1_S$ and $\mathcal{S}^1_T$, and $\mathcal{S}^2_S$ and $\mathcal{S}^2_T$.
The above scheme ensures that the app labels in $\mathcal{S}^2$ were indeed never encountered while the memory network was being trained on $\mathcal{S}^1$, and are hence unseen. It also ensures that test HLIs were never encountered during training, and are not a part of the support set either.

To facilitate learning and prediction, we ensure that for every app $a_l\in L_1$, there exists some $(h_1^1, a_l)\in\mathcal{S}^1_S,(h_2^1, a_l)\in \mathcal{S}^1_T$. Similarly for $L^2,\mathcal{S}^2_S,\mathcal{S}^2_T$.
This ensures that there is some example in the support set that the memory network / word2vec neighbour model can use to assign the correct unseen label to a test case. 

$L_1$ and $L_2$ were selected so as to meet all the above requirements.
Without discarding a few labels as discussed above, this would not have been possible.
With all the above conditions met, $L^1$ consisted of 59 apps, while $L^2$ consisted of 62 apps.
Each app in $L^2$ featured in only one example in each of $\mathcal{S}^2_S$ and $\mathcal{S}^2_T$, making it possible to evaluate the models in a 1-shot learning task setting. A vast majority of apps in $L^1$ met this condition too, with respect to $\mathcal{S}^1_S$ and $\mathcal{S}^1_T$.

In a top-n setting, where the model returns a list of top n labels from a ranked list, we consider a test sample as correctly classified if the correct app label is present in the top n labels.
We compare the performance of two models on $\mathcal{S}^2_T$.
One is the word2vec nearest neighbour model (see \ref{sec:FFNN}) with $\mathcal{S}^2_S$ as its support set.
The other is the best performing memory network architecture (see \ref{sec:MatNet}), trained using $\mathcal{S}^1_S$ as the support set and $\mathcal{S}^1_T$ as the training target; and tested on $\mathcal{S}^2_T$ with $\mathcal{S}^2_S$ as the support set.
The accuracy of the classification task is described in Table \ref{tbl:unseen_app_perf}.

\begin{table}[ht]
\centering
\caption{Performance of models on unseen apps in a 1-shot learning scenario}
\label{tbl:unseen_app_perf}
\resizebox{\textwidth}{!}{
\begin{tabular}{|c|c|c|c|c|c|c|c|c|c|c|}
\hline
&top-1       & top-2 & top-3 & top-4 & top-5 & top-6 & top-7 & top-8 & top-9 & top-10      \\ \hline
word2vec NN & 20.51 & 44.87 & 51.28 & 51.28 & 51.28 & 52.56 & 52.56 & 53.85 & 56.41  & 57.69 \\ \hline
Mem Net     & 26.92 & 44.87 & 50.00 & 51.28 & 51.28 & 51.28 & 51.28 & 51.28 & 52.56  & 55.13 \\ \hline
\end{tabular}}
\end{table}

We observe that the matching network yields a higher top-1 purity. Otherwise, the performance if comparable to that of the word2vec nearest neighbour model.
Ultimately, it will depend upon the task and the amount of  training data available, when deciding whether the gain in performance is worth the extra training data needed by the memory network.

\section{Conclusion}

We thus present results from a variety of approaches for multi-label classification on our dataset.
We find that simpler matching network architectures are able to work better than other approaches when they are trained on a dataset that resembles the test set, even in a limited sized dataset with only a handful examples of each class.
The model still performs better, but to a much less degree, in the 1-shot learning setting, in comparison to the word2vec nearest neighbour model, which has the advantage of not requiring any extra training data.
We observe that given the limited data, certain memory network architectures are able to perform better than others.
Overall, we conclude that memory networks remain a very viable option, even for tasks requiring classification over a large number of classes, when the training data is limited.

\bibliographystyle{sbc}
\bibliography{biblio}

\begin{thebibliography}{}

\bibitem[Kiros et~al. 2015]{Kiros2015SkipThoughtV}
Kiros, R., Zhu, Y., Salakhutdinov, R., Zemel, R.~S., Urtasun, R., Torralba, A.,
  and Fidler, S. (2015).
\newblock Skip-thought vectors.
\newblock In {\em NIPS}.

\bibitem[Mikolov et~al. 2013]{Mikolov2013EfficientEO}
Mikolov, T., Chen, K., Corrado, G.~S., and Dean, J. (2013).
\newblock Efficient estimation of word representations in vector space.
\newblock {\em CoRR}, abs/1301.3781.

\bibitem[Mueller and Thyagarajan 2016]{Mueller2016SiameseRA}
Mueller, J. and Thyagarajan, A. (2016).
\newblock Siamese recurrent architectures for learning sentence similarity.
\newblock In {\em AAAI}.

\bibitem[Sukhbaatar et~al. 2015]{Sukhbaatar2015EndToEndMN}
Sukhbaatar, S., Szlam, A., Weston, J., and Fergus, R. (2015).
\newblock End-to-end memory networks.
\newblock In {\em NIPS}.

\bibitem[Sun 2016]{Sun2016AdaptingSD}
Sun, M. (2016).
\newblock Adapting spoken dialog systems towards domains and users.

\bibitem[Sun et~al. 2016a]{Sun2016AnIA}
Sun, M., Chen, Y.-N., and Rudnicky, A.~I. (2016a).
\newblock An intelligent assistant for high-level task understanding.
\newblock In {\em IUI}.

\bibitem[Sun et~al. 2016b]{Sun2016WeaklySU}
Sun, M., Pappu, A., Chen, Y.-N., and Rudnicky, A.~I. (2016b).
\newblock Weakly supervised user intent detection for multi-domain dialogues.
\newblock In {\em SLT}.

\bibitem[Vinyals et~al. 2016]{Vinyals2016MatchingNF}
Vinyals, O., Blundell, C., Lillicrap, T., Kavukcuoglu, K., and Wierstra, D.
  (2016).
\newblock Matching networks for one shot learning.
\newblock In {\em NIPS}.

\end{thebibliography}

\end{document}